\documentclass[lettersize,journal]{IEEEtran}
\usepackage{amsmath,amsfonts}
\usepackage{algorithmic}
\usepackage{algorithm}
\usepackage{array}
\usepackage{textcomp}
\usepackage{stfloats}
\usepackage{url}
\usepackage{verbatim}
\usepackage{graphicx}
\usepackage{cite}
\usepackage[colorlinks,
            linkcolor=red,
            anchorcolor=blue,
            citecolor=green
            ]{hyperref}
\usepackage{authblk}
\usepackage{graphicx}
\usepackage{amssymb,amsmath,epsfig}
\usepackage{algorithm}
\usepackage{algorithmic}
\usepackage{booktabs}
\usepackage{mathrsfs}
\usepackage{color}
\usepackage{textcomp}
\usepackage{bbding}
\usepackage{pifont}
\usepackage{wasysym}
\usepackage{amssymb}
\usepackage{subcaption}
\usepackage{cite}
\usepackage{amsmath,amssymb,amsfonts}
\usepackage{xcolor}
\usepackage{multirow}
\usepackage{epsfig}
\usepackage{colortbl}
\usepackage{graphicx}
\usepackage{multirow}
\usepackage{colortbl}
\usepackage{xcolor}
\usepackage{array}
\usepackage{bbding}
\usepackage{array}

\usepackage{booktabs}

\begin{document}

\title{MacFormer: Semantic Segmentation with \\ Fine Object Boundaries}

\author{
Guoan Xu, 
Wenfeng Huang, 
Tao Wu,
Ligeng Chen,
Wenjing Jia,~\IEEEmembership{Member,~IEEE} 
\\
Guangwei Gao,~\IEEEmembership{Senior Member,~IEEE}
Xiatian Zhu,
Stuart Perry
\thanks{Guoan Xu, Wenfeng Huang, Wenjing Jia, and Stuart Perry are with the Faculty of Engineering and Information Technology, University of Technology Sydney, Sydney, Australia (e-mail: xga\_njupt@163.com, huang-wenfeng@outlook.com, Wenjing.Jia@uts.edu.au, and Stuart.Perry@uts.edu.au).}

\thanks{Tao Wu and Ligeng Chen are with the Faculty of Computer Science, Nanjing University, Nanjing, China (e-mail: wt@smail.nju.edu.cn, and  chenlg@smail.nju.edu.cn).}

\thanks{Xiatian Zhu is with the Surrey Institute of People-Centred AI, and Centre for Vision, Speech and Signal Processing (CVSSP), Faculty of Engineering and Physical Sciences, University of Surrey, Guildford, United Kingdom (e-mail: Eddy.zhuxt@gmail.com).}

\thanks{Guangwei Gao is with the Institute of Advanced Technology, Nanjing University of Posts and Telecommunications, Nanjing, China, and also with the Key Laboratory of Artificial Intelligence, Ministry of Education, Shanghai, China (e-mail: csggao@gmail.com).}

}

\markboth{Journal of \LaTeX\ Class Files,~Vol.~14, No.~8, August~2021}%
{Shell \MakeLowercase{\textit{et al.}}: A Sample Article Using IEEEtran.cls for IEEE Journals}


\maketitle

\begin{abstract}

Semantic segmentation involves assigning a specific category to each pixel in an image. While Vision Transformer-based models have made significant progress, current semantic segmentation methods often struggle with precise predictions in localized areas like object boundaries. To tackle this challenge, we introduce a new semantic segmentation architecture, ``MacFormer'', which features two key components. Firstly, using learnable agent tokens, a Mutual Agent Cross-Attention (MACA) mechanism effectively facilitates the bidirectional integration of features across encoder and decoder layers. This enables better preservation of low-level features, such as elementary edges, during decoding. Secondly, a Frequency Enhancement Module (FEM) in the decoder leverages high-frequency and low-frequency components to boost features in the frequency domain, benefiting object boundaries with minimal computational complexity increase. MacFormer is demonstrated to be compatible with various network architectures and outperforms existing methods in both accuracy and efficiency on benchmark datasets ADE20K and Cityscapes under different computational constraints.
\end{abstract}


\section{Introduction}
\IEEEPARstart{S}{emantic} segmentation, a crucial task in computer vision, has been a prominent research focus~\cite{long2015fully,wang2020deep,zheng2021rethinking}. It entails assigning a specific category to each pixel in an image, making it a dense prediction task rather than a prediction at the image level. This task finds applications across a wide range of real-world scenarios~\cite{asgari2021deep,gao2021mscfnet,yuan2021review}. 

Recently, the emergence of Vision Transformers (ViT)~\cite{dosovitskiy2020image} has ignited a significant revolution in the field, as initially explored by Zheng \textit{et al.}~\cite{zheng2021rethinking}. While ViT shows promising performance, it has some limitations: 1) ViT generates single-scale low-resolution features instead of multi-scale ones. 2) It involves high computational overhead, particularly for large images. To overcome these challenges, Wang \textit{et al.}~\cite{wang2021pyramid} introduced a Pyramid Vision Transformer (PVT) as an extension of ViT, incorporating pyramid structures for dense prediction. However, newer approaches like Swin Transformer~\cite{liu2021swin} and Twins~\cite{chu2021twins} primarily focus on refining the Transformer encoder and may overlook the potential enhancements from the decoder side. 
\begin{figure}[t]
    \centering
    \begin{subfigure}[b]{0.23\textwidth}
        \centering
        \includegraphics[width=\textwidth]{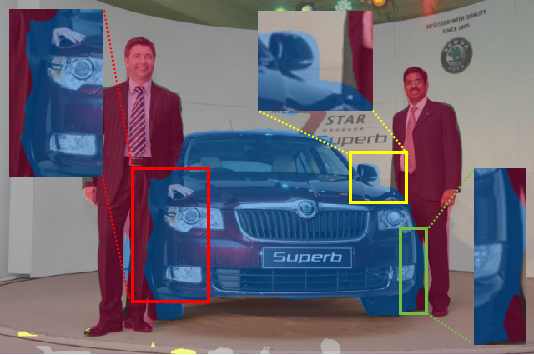}
        \caption{SegFormer~\cite{xie2021segformer}}
        \label{fig:SegFormer}
    \end{subfigure}
    \hfill
    \begin{subfigure}[b]{0.23\textwidth}
        \centering
        \includegraphics[width=\textwidth]{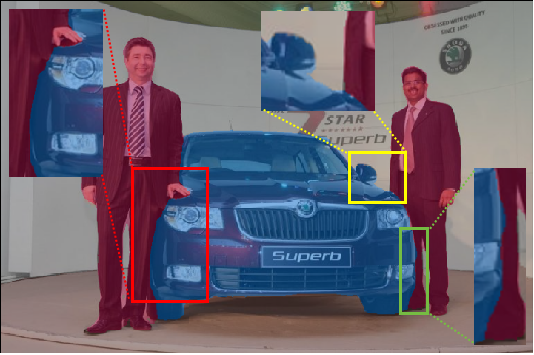}
        \caption{MacFormer (ours)}
        \label{fig:MacFormer}
    \end{subfigure}
    \caption{Addressing the challenge of accurate segmentation at object boundaries, which often faces interference from neighboring categories, our MacFormer offers a promising solution. Visual comparison to SegFormer~\cite{xie2021segformer} showcases our method's superior performance in segmenting detailed features, particularly at the edges of individuals and vehicles.} 
    \label{Figure 0}
\end{figure}

Other studies have delved into incorporating transformer decoder structures into vision tasks. The pioneering work of DETR~\cite{carion2020end} introduced the transformer encoder-decoder architecture to tasks such as 
detection and segmentation. Following this, Segmenter~\cite{strudel2021segmenter}, MaskFormer~\cite{cheng2021per}, and Mask2Former~\cite{cheng2022masked} integrated decoders for mask prediction with global class labels, emphasizing the use of high-level features. However, conventional transformer decoders typically utilize class-wise queries to extract feature information for each class, often relying on the highest-level feature or requiring an additional decoder for multi-scale features. This process involves decoding class-wise queries for each multi-scale feature, significantly increasing computational complexity.

Recently, FeedFormer~\cite{shim2023feedformer} introduced an innovative decoder architecture that leverages the lowest-level encoder feature to decode high-level encoder features. Additionally, U-MixFormer~\cite{yeom2023u} presented a unique mix-attention mechanism that dynamically incorporates multi-stage features as keys and values. This specialized module enables the gradual transmission of features across decoder stages, iteratively remixing them to address dependencies and enhance contextual understanding. Both approaches employ cross-attention within the decoder to facilitate information flow from the encoder, enabling the integration of multi-scale features. However, they struggle 
in accurately segmenting object boundaries and providing adequate unidirectional supplementation, lacking bidirectional information exchange and feature complementation.

Building on this, we introduce a novel attention mechanism, named ``mutual agent cross-attention''. This module enables the decoder to query encoder features and allows the encoder to obtain complementary information from the decoder in a multi-scale fashion. Each component queries the necessary keys and values, fostering mutual complementarity and resulting in more robust features. 
Inspired by the success of 
agent attention~\cite{han2023agent} on improving 
computational efficiency, we introduce additional agent tokens to serve as intermediaries, optimizing the computation process. The dimension of agent tokens can be adjusted to control the computational complexity of self-attention, leading to the efficient mutual agent cross-attention mechanism.

However, another critical issue in segmentation arises from the conflict between network depth and detail preservation. As the network becomes deeper, it excels in extracting abstract semantic context but tends to lose low-level details. 
Effectively preserving the structural details lost in the encoder remains a challenge. Drawing inspiration from prior studies~\cite{bar2003cortical,chen2019drop,kauffmann2014neural,si2022inception,bullier2001integrated}, it is noted that detailed texture information predominantly resides in high-frequency components, while low-frequency components contain rich global information. In this study, we propose transforming encoder features into the frequency domain to facilitate the separation of detailed information of high and low-frequency components. 
We then extract high-frequency components from shallow features to complement those separated from deep features, enriching the structural details. Simultaneously, the low-frequency components from deep features enhance those of shallow features, complementing deficiencies in both shallow and deep features. When compared to the state-of-the-art SegFormer method~\cite{xie2021segformer}, Fig.~\ref{Figure 0} illustrates our method's superior handling of details, particularly evident in the boundaries of the person and the car. 

The contributions of this paper can be summarized 
as follows:

\begin{enumerate}
\item 
We introduce a carefully crafted Mutual Agent Cross-Attention (MACA) mechanism to address the challenges of long-distance modeling and bidirectional feature supplementation in semantic segmentation. 

\item 
We propose a Frequency Enhancement Module (FEM) to preserve both the shallow-level object structures and deep semantic information. 
 
\item 
Extensive experiments are conducted using different backbones on 
benchmark datasets Cityscapes and ADE20K, demonstrating 
state-of-the-art performance. 
\end{enumerate}

\begin{figure*}[htbp]
	\centerline{\includegraphics[width=18cm]{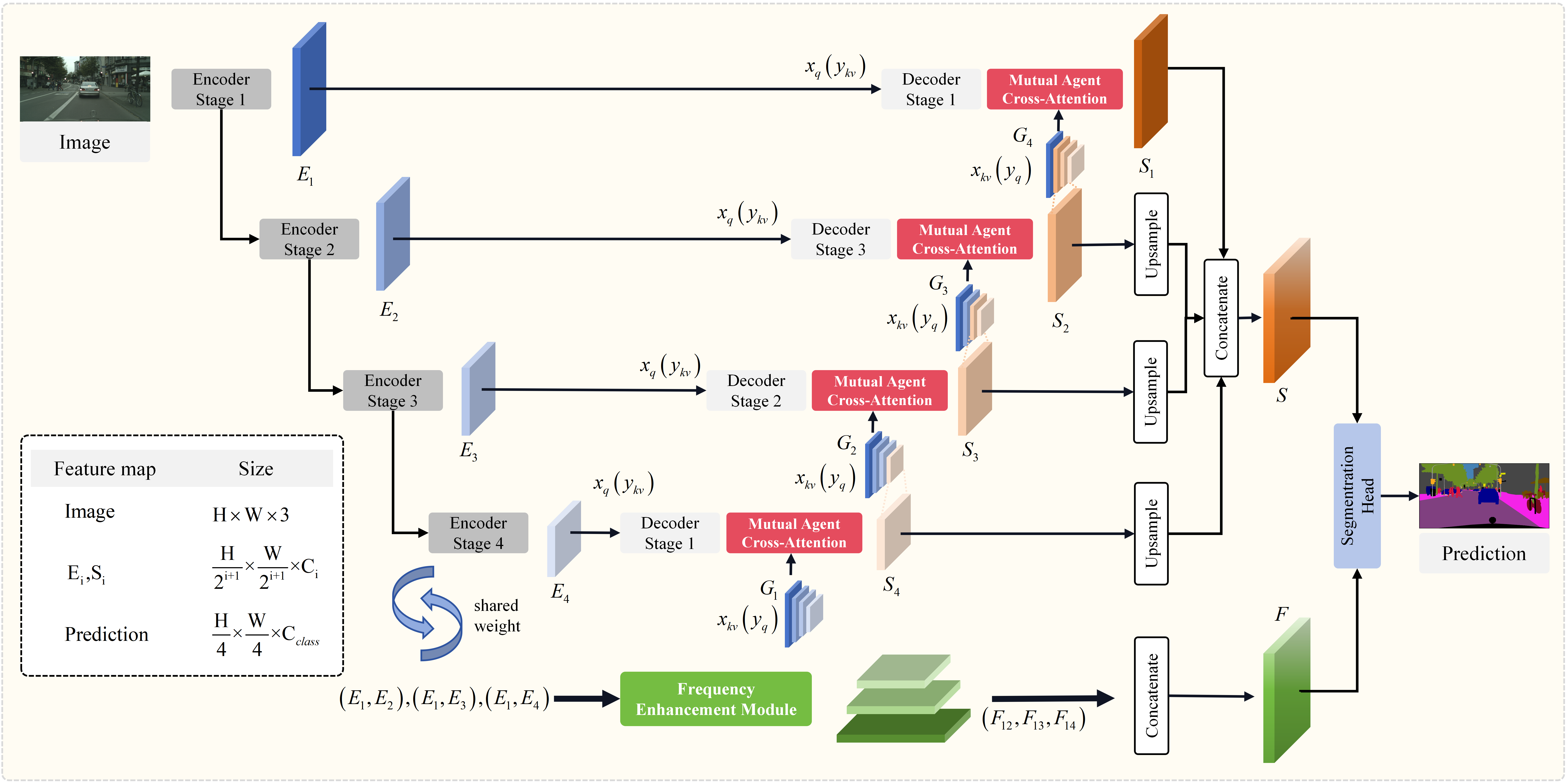}}
	\caption{The architecture of our proposed MacFormer consists of several key components. Initially, multi-scale feature maps are obtained in the Encoder. Subsequently, the Decoder incorporates the Mutual Agent Cross-Attention (MACA) mechanism and Frequency Enhancement Module (FEM) to improve the encoder features. Finally, the enhanced features are aggregated and forwarded to the segmentation head to produce the ultimate prediction.}
	\label{Figure 1}
\end{figure*}

\section{Related Work}
\label{sec2}

\subsection{Encoders in Semantic Segmentation} 
\label{sec22}
SETR~\cite{zheng2021rethinking}, introduced by Zheng \textit{et al.} in 2021, marked a pioneering step by utilizing the Vision Transformer (ViT) as its encoder for semantic segmentation tasks. 
Then, the research community explored various impactful enhancements, including the incorporation of multi-scale feature hierarchies and the integration of convolutional techniques. MViT~\cite{fan2021multiscale} is recognized for creating a multi-scale pyramid of features.
PVT~\cite{wang2021pyramid, wang2022pvt} introduced pyramid structures using an iterative process to group feature maps into distinct, non-overlapping patches at different encoder stages, leading to the hierarchical generation of multi-scale encoder features. 

On the other hand, Swin Transformer~\cite{liu2021swin} introduced an innovative hierarchical Transformer architecture that computes representations using shifted windows. 
HRFormer~\cite{yuan2021hrformer}, inspired by HRNet~\cite{wang2020deep}, took a unique approach by enhancing local features and enabling information exchange between non-overlapping windows. 
SegFormer~\cite{xie2021segformer} introduced
a Mix Transformer encoder and an ALL-MLP decoder in a unified structure. 
SegNeXt~\cite{guo2022segnext} 
employed a convolutional attention mechanism and combined successful segmentation model attributes, emphasizing a robust backbone, multi-scale interaction, and spatial attention for effectiveness. Additionally, techniques like CvT~\cite{wu2021cvt}, CoaT~\cite{xu2021co}, LeViT~\cite{graham2021levit}, and Twins~\cite{chu2021twins} enhanced feature local continuity and removed fixed-size position embeddings, improving Transformer performance in dense prediction tasks. Notably, for a balance of performance and efficiency, researchers commonly choose the MiT-B0$\sim$B5 series in Segformer~\cite{xie2021segformer} and MSCAN-T/S/B in SegNext~\cite{guo2022segnext}. 

Following this trend, in our work, we utilize effective hierarchical vision transformers as the backbone. 
It is worth noting that, our approach is backbone-generic, compatible with various network architectures.

\subsection{Decoders in Semantic Segmentation} 
\label{sec23}

Recent studies have utilized the Transformer decoder to enhance the performance of semantic segmentation. DETR~\cite{carion2020end} was a pioneering approach that employed a learned set of object queries to infer object-context relationships, resulting in a parallel set of predictions. Subsequent studies~\cite{strudel2021segmenter,xie2021segformer,cheng2022masked} have also adopted this scheme, integrating it with multi-scale encoder features extracted from the backbone, albeit leading to increased computational demands. 

For an efficient and lightweight decoder head, SegNext~\cite{guo2022segnext} used an MLP structure aggregating features from the final three stages and employed a streamlined Hamburger module to enhance global context modeling. 
In contrast, FeedFormer~\cite{shim2023feedformer} directly integrated encoder stage features as feature queries, with the lowest-level encoder features serving as keys and values, resulting in improved efficiency. 
To facilitate incremental feature propagation in the decoder, U-mixFormer~\cite{yeom2023u} drew inspiration from the UNet architecture 
and systematically incorporated mix-attention across different scale features layer by layer. 
While this operation is highly effective, there is still potential for further improvement. 



\subsection{Efficient Attention Mechanisms}
\label{sec24}

Efficiency is a bottleneck with Transformer's attention mechanism. Various methodologies~\cite{dong2022cswin,katharopoulos2020transformers} have emerged to tackle the GPU memory overheads linked to spatial self-attention. Axial attention~\cite{huang2022channelized,hou2020strip,wang2020axial} was devised to alleviate the computational load of traditional global self-attention by conducting self-attention along a single axis at a time and integrating horizontal and vertical axial attention modules to achieve a comprehensive global receptive field. GRL~\cite{li2023efficient} utilized a blend of anchored stripe attention, window attention, and channel attention to effectively perform image restoration.
However, these methods have a relative disadvantage in achieving optimal final results as they inherently limit the global receptive field of self-attention. 
Recently, Han \textit{et al.}~\cite{han2023agent} introduced 
a new agent attention mechanism that cleverly combines Softmax and linear attention. this approach introduces an agent token A alongside Q, K, and V, effectively balancing performance and computational complexity. 
Building upon this concept, we propose integrating the benefits of both attention mechanisms and cross-attending features from different stages, leading to a mutual agent cross-attention mechanism, as detailed in Section~~\ref{sec32}. 

\subsection{Fourier Frequency Domain}
\label{sec25}
Recent research~\cite{jiang2021focal,rao2021global,yu2022deep} has leveraged Fourier Transform on images, harnessing frequency information to enhance model performance and efficiency. 
It surpasses traditional spatial domain methods in capturing intricate geometric structures that are difficult to extract. These studies suggest that high-frequency components in the frequency domain contain detailed structural information, while low-frequency components are closely linked to high-level semantic information. 

SGNet~\cite{wang2023sgnet} employed an iterative approach that leverages real image frequency components to steer depth super-resolution of the depth map, yielding commendable outcomes. Rao \textit{et al.}~\cite{rao2021global} replaced Fast Fourier Transform (FFT) for self-attention modules in the traditional Transformer, effectively capturing global information while keeping computational overhead low. 

Meanwhile,~\cite{jiang2021focal} introduced a novel focal frequency loss tailored for Fourier spectrum supervision, enhancing the performance of prevalent image generative models.
FreMIM~\cite{wang2024fremim} proposed utilizing both the low-pass and high-pass Fourier spectrum separately as the supervision signal, along with a masking strategy.
Pham \textit{et al.}~\cite{pham2024frequency} skillfully integrated frequency information with knowledge distillation, leveraging the fact that each frequency is determined by all image pixels in the spatial domain. Utilizing frequency domain information to guide features in segmentation tasks is particularly significant, addressing challenges in recovering fine object boundary details during the upsampling process. Our proposed approach focuses on employing high-frequency information from shallow features to compensate for high-frequency components, aiming to achieve precise segmentation at object boundary areas, with promising results.

Building upon 
the above observations and insights, a persistent challenge in semantic segmentation lies in accurately predicting object boundaries. To tackle this issue, we propose 
leveraging both the attention mechanism and frequency domain information, as detailed below.

\section{Method}

Fig.~\ref{Figure 1} illustrates the detailed architecture of our proposed MacFormer, which is primarily composed of an encoder, a decoder strengthened with our Mutual Agent Cross-Attention (MACA) and Frequency Enhancement Module (FEM).

\subsection{The Overall Network Architecture}
\label{sec31}

Following the latest trend, we employ recently popular and effective hierarchical vision transformer backbones for the encoder: LVT~\cite{yang2022lite}, MiT from Segformer~\cite{xie2021segformer}, and MSCAN from SegNext~\cite{guo2022segnext}. 
Specifically, 
given an input image $X \in {R^{H \times W \times 3}}$, where $H$ and $W$ are the height and width of the input image, after processing with the backbones equipped with $i \in \left\{ {1,2,...,I|I = 4} \right\}$ stages of the encoder, we obtain multi-scale features, labeled as $E_i$ with resolutions $\frac{H}{{{2^{i + 1}}}} \times \frac{W}{{{2^{i + 1}}}} \times {C_i}$, where $C_i$ is the channel of the feature map and then feed them into the decoder.

Each stage of the encoder extracts features with distinct resolutions. 
Features extracted from earlier stages encompass abundant boundary details, whereas those from deeper stages harbor higher-level semantic understanding. How to make the most of these features, which encapsulate diverse information, is critical for effective image segmentation. 
Inspired by~\cite{shim2023feedformer,zhang2022topformer,yeom2023u}, we adopt the U-shaped structure, which is better able to propagate features from the encoder to the decoder, thereby guiding the restoration of feature resolution. 

Our decoder mainly consists of two modules:  Mutual Agent Cross Attention (MACA) and Frequency Enhancement Modules (FEM). 
It iteratively refines features $E_i$ from the encoder through the 
MACA module across multiple stages and obtains features ${S_{i}}$. Subsequently, the features ${S_{i}}$ undergo upsampling to match the height and width of $E_1$. The features processed by the MACA module are concatenated to obtain the spatial enhanced feature $S$. 

While utilizing the attention mechanism enhances pixel-level correlations in existing spatial domain features, it fails to sufficiently compensate for the information lost during downsampling in the encoder. Hence, we come up with utilizing features with richer object boundary details in the frequency domain to address this issue.

\begin{figure}[tbp]
	\centerline{\includegraphics[width=8.5cm]{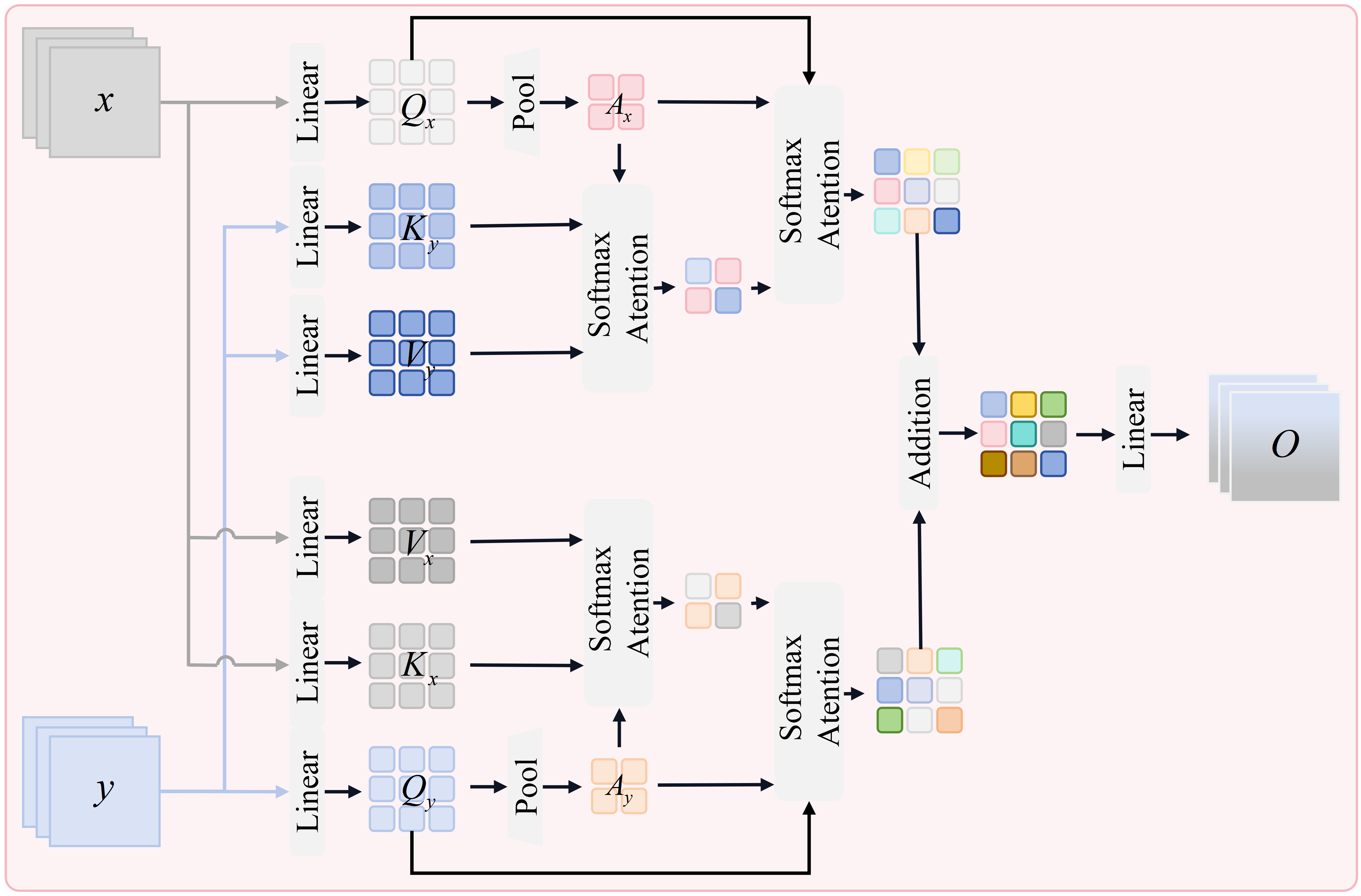}}
	\caption{The proposed Mutual Agent Cross-Attention (MACA) mechanism involves cross-operations among tokens from various features, along with tokens represented as $A$, resulting in enhanced performance. The computational complexity can be managed by adjusting the $A$ dimension.}
	\label{Figure 2}
\end{figure}

In general, 
images' 
structural information primarily resides in their high-frequency components, while high-level semantics are abundant in their low-frequency counterparts~\cite{wang2020high,si2022inception}. Leveraging this property, we propose to employ the high-frequency information from the shallow feature $E_1$ to compensate the high-frequency components of $E_i$ $(i=2,3,4)$ and hence enhance the features. 

Specifically, this involves subtracting the high-frequency components of $E_1$ from those of $E_i$ $(i=2,3,4)$, then concatenating the frequency difference to the original frequency components of $E_i$. 

Consequently, the edge details in $E_i$ $(i=2,3,4)$ receive significant complementation, 
and the low-frequency components in $E_i$ $(i=2,3,4)$, which contain intricate semantic information, complement 
$E_1$ at the same time. 
Fig.~\ref{Figure 5} illustrates the benefits brought by the frequency enhancement module. 

The features processed by the FEM, 
are then concatenated to obtain the enhanced feature $F$. Eventually, the spatial-enhanced feature $S$ and the frequency-enhanced feature $F$ are summed together. After processing by a segmentation head, the final segmentation map is obtained. 
Section~\ref{sec33} presents the details of the FEM process. 

\subsection{Mutual Agent Cross-Attention (MACA)}
\label{sec32}

To address the computational complexity issue of self-attention, many researchers have proposed their own solutions~\cite{liu2023efficientvit,katharopoulos2020transformers}. 
One recent popular method is Agent-attention~\cite{han2023agent}. In this approach, agent tokens $A$ are first treated as queries, initiating attention computations between $A$, $K$, and $V$ to gather agent features $V_A$ from all values. Then, $A$ is utilized as keys and $V_A$ as values in a subsequent attention calculation alongside the query matrix $Q$. This step broadcasts global information from agent features to every query token, resulting in the final output $\rm O^{Agent}$. Agent attention can be formulated as follows:

\begin{equation}
    {{\mathop{\rm O^{Agent}}\nolimits}} = Softmax\left( {\frac{{Q{A^T}}}{{\sqrt {{d_{head}}} }}} \right)Softmax\left( {\frac{{A{K^T}}}{{\sqrt {{d_{head}}} }}} \right)V,
\end{equation}
where $A \in {R^{n \times d}}$ is the newly defined agent tokens.

The distinction between cross-attention and self-attention lies in that, the features used to generate queries, keys, and values are identical ($X_{qkv}$) and originate from the same source in self-attention, while cross-attention employs two distinct features, $X_q$ and $X_{kv}$, each derived from a single unique source. In this way, $X_q$ can index the features of the $X_{kv}$ sequence, thus inheriting its outstanding characteristics. If we denote two features as $x$ and $y$, and project them to obtain tokens $Q_x$, $K_x$, $V_x$ and $Q_y$, $K_y$, $V_y$ respectively, the calculation process of cross attention is as follows:
\begin{equation}
    {{\mathop{\rm O}\nolimits} ^{Cross}} = Softmax\left( {\frac{{{Q_x}{K_y}^T}}{{\sqrt {{d_{head}}} }}} \right){V_y}.
\end{equation}

\begin{figure*}[tbp]
	\centerline{\includegraphics[width=17cm]{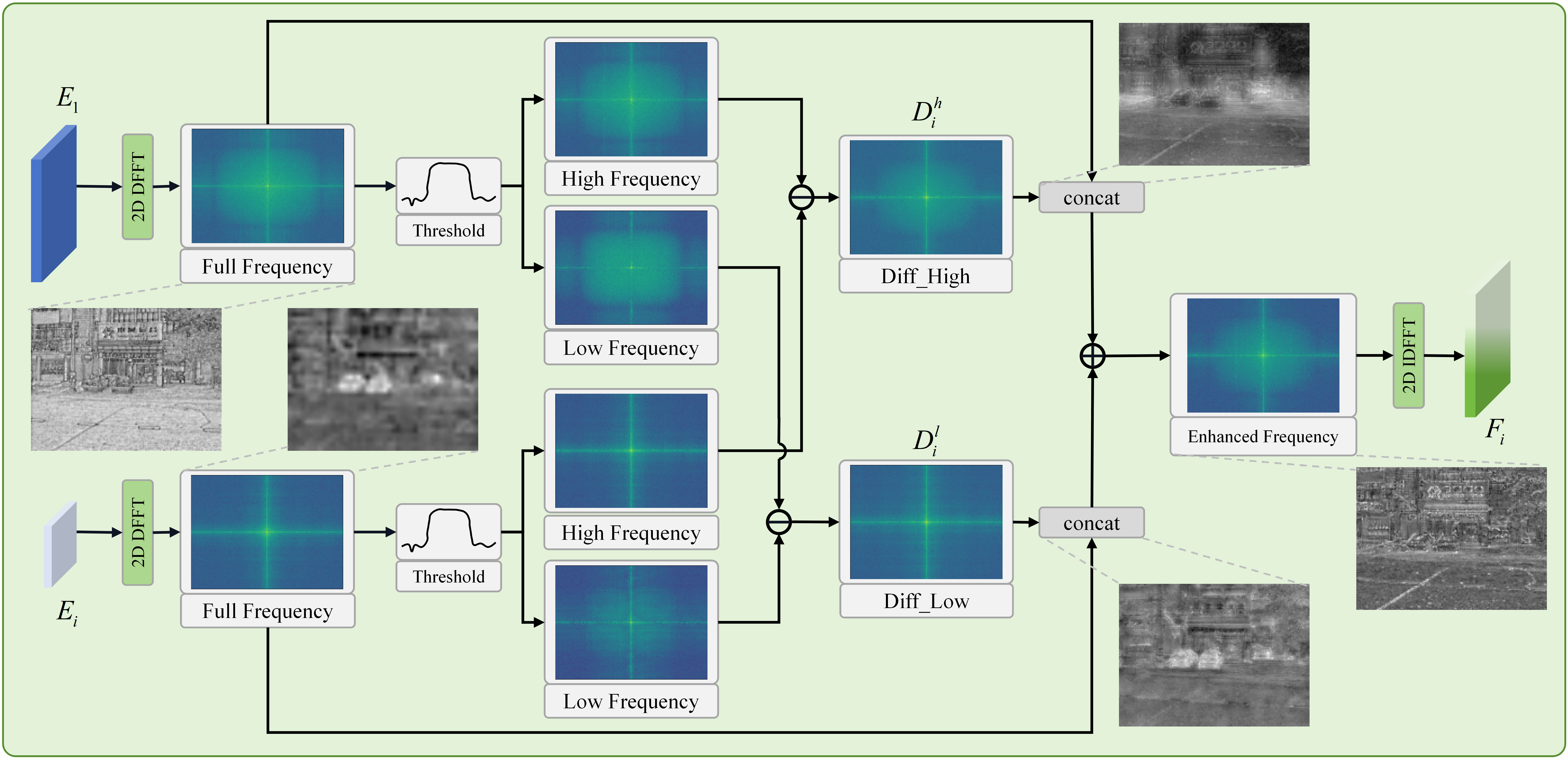}}
	\caption{The illustration of the proposed Frequency Enhancement Module (FEM). The advantage of $E_1$ lies in its capability of capturing 
 detailed high-frequency information, while $E_i$ excels at retaining low-frequency semantic context.}
	\label{Figure 3}
\end{figure*}

Cross-attention enables the feature $x$ to obtain information from feature $y$, but overlooks the fact that feature $y$ also desires information from feature $x$. 
We propose to integrate the advantages of the two types of attention mechanisms 
and cross-attending features from different stages to leverage the strengths of each other, resulting in mutual agent cross-attention shown in Fig.~\ref{Figure 2}. 

The formula for mutual cross-attention is presented as:
\begin{equation}
{{\mathop{\rm O}\nolimits} ^{m\_Cross}} = {\mathop{\rm O}\nolimits}_{x \to y}^{Cross} + {\mathop{\rm O}\nolimits}_{y \to x}^{Cross},
\end{equation}
where
\begin{equation}
    {\mathop{\rm O}\nolimits} _{x \to y}^{Cross} = Softmax\left( {\frac{{{Q_x}K_y^T}}{{\sqrt {{d_{head}}} }}} \right){V_y},
\end{equation}
\begin{equation}
    {\mathop{\rm O}\nolimits} _{y \to x}^{Cross} = Softmax\left( {\frac{{{Q_y}K_x^T}}{{\sqrt {{d_{head}}} }}} \right){V_x}.
\end{equation}

However, directly adopting mutual cross-attention inevitably brings about a heavier computational burden. Therefore, by incorporating the concept of agent attention, in this paper we propose to merge agent attention and cross attention and develop a new Mutual Agent Cross-Attention (MACA) mechanism, 
which can be expressed as follows:
\begin{equation}
    {{\mathop{\rm O}\nolimits} ^{mA\_Cross}} = {\mathop{\rm O}\nolimits} _{x \to y}^{mA\_Cross} + {\mathop{\rm O}\nolimits} _{y \to x}^{mA\_Cross},
\end{equation}
where 
\begin{equation}
{\mathop{\rm O}\nolimits} _{x \to y}^{mA\_Cross} = \sigma \left( {\frac{{{Q_x}{A^T}}}{{\sqrt {{d_{head}}} }}} \right) \sigma \left( {\frac{{AK_y^T}}{{\sqrt {{d_{head}}} }}} \right){V_y},
\end{equation}

\begin{equation}
    {\mathop{\rm O}\nolimits} _{y \to x}^{mA\_Cross} = \sigma \left( {\frac{{{Q_y}{A^T}}}{{\sqrt {{d_{head}}} }}} \right)\sigma \left( {\frac{{AK_x^T}}{{\sqrt {{d_{head}}} }}} \right){V_x}.
\end{equation}
Here, $\sigma$ represents the $Softmax$ function.

\begin{figure*}[htbp]
  \centering
  \scalebox{0.95}{
  \begin{subfigure}[b]{0.22\textwidth}
    \centering
    \includegraphics[width=\textwidth]{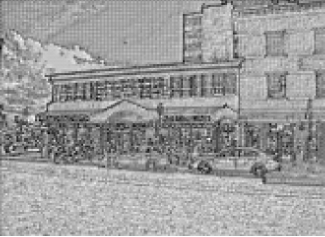}
    \caption{$E_1$}
    \label{fig:image1}
  \end{subfigure}
  \begin{subfigure}[b]{0.22\textwidth}
    \centering
    \includegraphics[width=\textwidth]{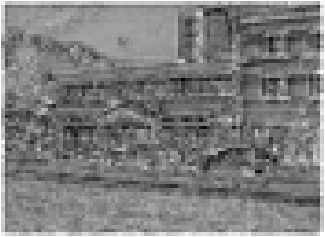}
    \caption{$E_2$}
    \label{fig:image2}
  \end{subfigure}
  \begin{subfigure}[b]{0.22\textwidth}
    \centering
    \includegraphics[width=\textwidth]{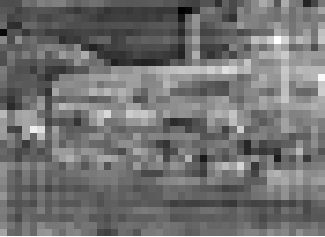}
    \caption{$E_3$}
    \label{fig:image3}
  \end{subfigure}
  \begin{subfigure}[b]{0.22\textwidth}
    \centering
    \includegraphics[width=\textwidth]{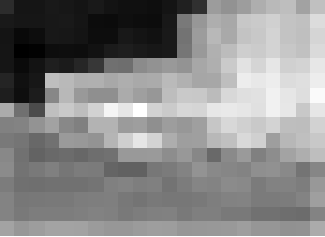}
    \caption{$E_4$}
    \label{fig:image4}
  \end{subfigure}
  }
  \\
  \scalebox{0.95}{
  \begin{subfigure}[b]{0.22\textwidth}
    \centering
    \includegraphics[width=\textwidth]{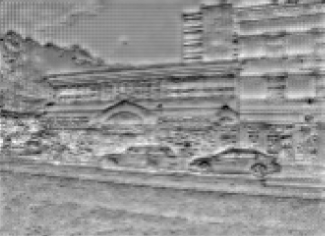}
    \caption{$F_{12}$}
    \label{fig:image5}
  \end{subfigure}
  \begin{subfigure}[b]{0.22\textwidth}
    \centering
    \includegraphics[width=\textwidth]{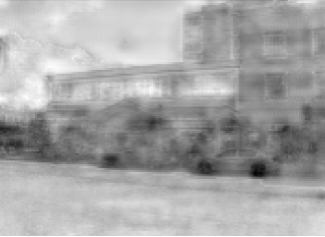}
    \caption{$F_{13}$}
    \label{fig:image6}
  \end{subfigure}
  \begin{subfigure}[b]{0.22\textwidth}
    \centering
    \includegraphics[width=\textwidth]{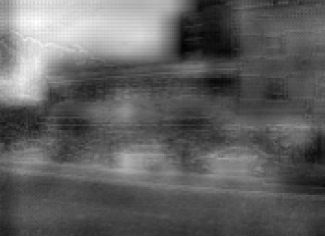}
    \caption{$F_{14}$}
    \label{fig:image7}
  \end{subfigure}
  \begin{subfigure}[b]{0.22\textwidth}
    \centering
    \includegraphics[width=\textwidth]{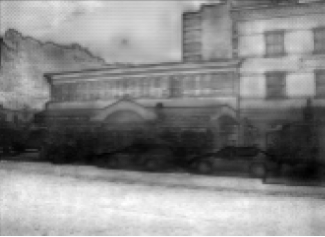}
    \caption{$F$}
    \label{fig:image8}
  \end{subfigure}
  }
  \caption{Visualization of feature maps. 
  The first row depicts the output feature maps from different stages of the Encoder. The second row illustrates the visualization of feature maps obtained after incorporating frequency domain information using FEM, where mutual supplementation can be observed. Through this supplementation, both detailed and semantic information are enhanced and fused. $F$ represents the result obtained after concatenating the three features, $F_{12}$, $F_{13}$ and $F_{14}$.}
  \label{Figure 4}
\end{figure*}

The formalization of selecting a feature set $G_j$ for a decoder stage $j$ in the spatial domain is expressed piecewise as follows:
\begin{equation}
    {G_j} = \left\{ \begin{array}{l}
\left\{ {{E_k}} \right\}_{k = 1}^I,\quad j = 1,\\
\left\{ {{E_k}} \right\}_{k = 1}^{I - j + 1} \cup \left\{ {{S_k}} \right\}_{k = I - j + 2}^I,\quad j \ne 1,
\end{array} \right.
\end{equation}
where $I$ is the number of encoder stages, 
set to $4$ in experiments. 
Thus, in the case of the first decoder stage $(j = 1)$, all encoder features are selected. Subsequent stages involve the propagation of previously computed decoder stage outputs by substituting them for their corresponding lateral encoder counterparts in $G_j$. 
Therefore, the enhanced features $S_i$ can be obtained by: 
\begin{equation}
    {S_i} = {{\mathop{\rm O}\nolimits} ^{mA\_Cross}}\left( {{E_i},{G_{I - i + 1}}} \right),\quad i = 1,...,I.
\end{equation}

Finally, $S_i$ are upsampled to the same size as $E_1$ and concatenated together, resulting in the final output feature S. 

The effectiveness of this module is demonstrated in our experiments presented in the Ablation Studies section.

\begin{figure*}[tbp]
	\centerline{\includegraphics[width=15cm]{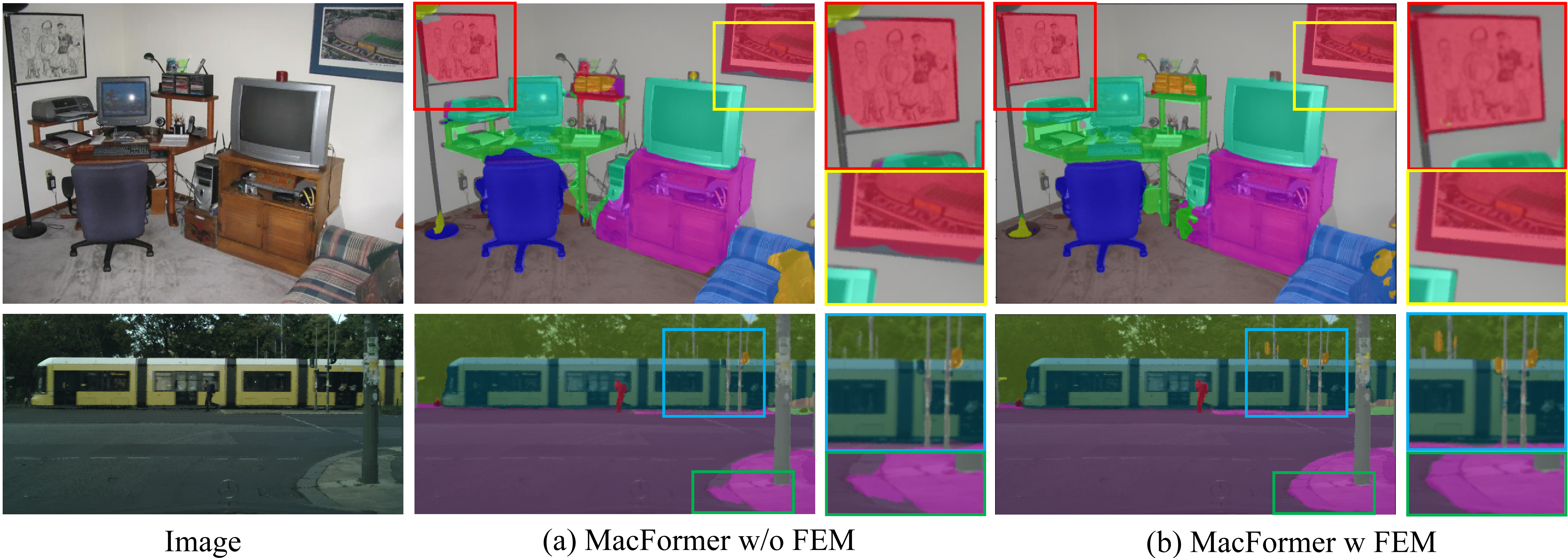}}
    \caption{Visual evaluation of our proposed FEM. 
It can be seen that after integrating the FEM, the segmentation at object boundaries are more precise. }
    \label{Figure 5}
\end{figure*}

\subsection{Frequency Enhancement Module (FEM)}
\label{sec33}

Widely recognized in image processing, the frequency domain excels at extracting object structure information. 


Specifically, 
consider a feature map $X$ with dimensions $H \times W \times C_{in}$. 
We first 
transform it into the frequency domain using Discrete Fast Fourier Transform (DFFT). 

For the $i$-th channel $X_i$ within the feature map $X$, its 2D DFT, denoted as $f_i$, 
is expressed as:
\begin{equation}
    {f_i}\left( {u,v} \right) = {f_{DFFT}}\left( {{X_i}} ( {h,w}) \right),
\end{equation}
where $X_i(h, w)$ represents the signal at coordinates $(h, w)$ in $X$, and $u$, $v$ are the indices corresponding to horizontal and vertical spatial frequencies in the Fourier spectrum. 

Similarly, the formulation for the 2D Inverse Discrete Fast Fourier Transform (2D-IDFFT) is as follows:
\begin{equation}
    X\left( {h,w} \right) = {f_{DFFT}}\left( {{f_i}} ( {u,v}) \right).
    \end{equation}

As demonstrated in prior studies~\cite{bar2003cortical,chen2019drop,kauffmann2014neural,si2022inception}, the intricate structural details within an image predominantly reside in the high-frequency components of the Fourier spectrum, whereas the comprehensive global semantic information is abundant in its low-frequency counterparts. During the decoding phase, object boundaries are lost in the deep features. However, these missing details can be complemented using the high-frequency components of shallow features. 

Therefore, as shown in Fig.~\ref{Figure 3}, in our Frequency Enhancement Module, we first apply the Fourier transform to encoder features $E_i$ as:
\begin{equation}
    E_i^{fre} = {f_{DFFT}}\left( {{E_i}} \right).
\end{equation}
Then, we set a threshold to extract the high-frequency and low-frequency components: 
\begin{equation}
    \begin{cases}
     H{i_{E_i^{fre}}} = E_i^{fre},\quad\left| {E_i^{fre}} \right| \ge threshold,\\
     L{o_{E_i^{fre}}} = E_i^{fre},\quad\left| {E_i^{fre}}  \right| < threshold,
    \end{cases}
\end{equation}
where $H{i_{E_i^{fre}}}$ and $L{o_{E_i^{fre}}}$ represent the high and low frequency information of frequency features $E_i^{fre}$, respectively. 

Subsequently, we utilize the high-frequency components of $E_1^{fre}$ to subtract the high-frequency components of $E_i^{fre}$ $(i=2,3,4)$. The difference between these two represents the structural details missing in the feature $E_i$ in the frequency domain: 
\begin{equation}
    {D_i^h} = H{i_{E_1^{fre}}} - H{i_{E_i^{fre}}}, \quad(i = 2, 3, 4).
\end{equation}

Then, we concatenate this difference $D_i^h$ with the original frequency information $E_i^{fre}$ $(i=2, 3, 4)$, resulting in the enhanced frequency domain feature $F_{1 \to i}^{fre}$ $(i=2, 3, 4)$, as:
\begin{equation}
    F_{1 \to i}^{fre} = Concat\left[ {D_i^h,{E_i^{fre}}} \right], \quad(i = 2, 3, 4).
\end{equation}

Meanwhile, 
we leverage the low-frequency components of $E_i$ $(i = 2, 3, 4)$ subtracted by the low-frequency component $E_1$. 
This difference signifies the high-level semantic information, represented by low-frequency components, lacking in $E_1$, as: 
\begin{equation}
    {D_i^l} = L{o_{E_i^{fre}}} - L{o_{E_1^{fre}}}, \quad(i = 2, 3, 4).
\end{equation}

Following this, we concatenate the difference ${D_i^l}$ with the original frequency information of $E_1^{fre}$, obtaining the enhanced $F_{i \to 1}^{fre}$ as:
\begin{equation}
    F_{i \to 1}^{fre} = Concat\left[ {D_i^l,{E_1^{fre}}} \right],\quad\left( {i = 2,3,4} \right).
\end{equation}
Then, the enhanced features $F_{1 \to i}^{fre}$ and $F_{i \to 1}^{fre}$ are added and processed through the inverse Fourier transform to obtain the ultimate enhanced features $F_i$, as:
\begin{equation}
    F_{1i} = IDFFT\left( {F_{i \to 1}^{fre} + F_{1 \to i}^{fre}} \right),\quad(i = 2,3,4).
\end{equation}

Finally, the features $F_{1i}$ are concatenated to obtain the ultimately enhanced feature $F$ in the frequency domain, 

Visualization of the intermediate feature maps can be seen in Fig.~\ref{Figure 4}, and the segmentation results obtained after merging frequency domain information using FEM are shown in Fig.~\ref{Figure 5}.

\begin{table*}[!t]
\caption{Performance comparison with SOTA light-weight models on ADE20K and Cityscapes.}
\begin{center}	
\scalebox{1.0}{
\begin{tabular}{llcccccc}
\toprule
\multirow{2}{*}{Method} & \multirow{2}{*}{Year} & \multirow{2}{*}{Backbone} & \multirow{2}{*}{Params. (M)$\downarrow$} & \multicolumn{2}{c}{ADE20K} & \multicolumn{2}{c}{Cityscapes} \\
&   &   &   & GFLOPs$\downarrow$ & mIoU (\%)$\uparrow$ & GFLOPs$\downarrow$ & mIoU (\%)$\uparrow$      \\ \midrule

FCN~\cite{long2015fully}   & 2015 CVPR  & MobileNet-V2  & 9.8  & 39.6  & 19.7  
& 317.1  & 61.5     \\
PSPNet~\cite{zhao2017pyramid}  & 2017 CVPR & MobileNet-V2 & 13.7  & 52.9  & 29.6   
& 423.4  & 70.2      \\
DeepLabV3+~\cite{chen2018encoder}  & 2018 ECCV & MobileNet-V2  & 15.4  & 69.4  
& 34.0   & 555.4  & 75.2   \\
SwiftNetRN~\cite{cheng2019swiftnet}  & 2019 arxiv   & ResNet-18   & 11.8   & -   & - & 104.0  & 75.5 \\
Semantic FPN~\cite{li2023convmlp}   & 2021 CVPR& ConvMLP-S & 12.8  & 33.8   
& 35.8   & -  & -  \\

\midrule

SegFormer~\cite{xie2021segformer}  & 2021 NeurIPS & MiT-B0 & 3.8 & 8.4  & 37.4   
& 125.5  & 76.2 \\
SeaFormer~\cite{wanseaformer}
& 2023 ICLR & SeaFormer-S 
& 4.0 & 1.1 & 38.1 & - & 76.1 \\
SCTNet~\cite{xu2024sctnet} 
& 2024 AAAI & SCTNet-S
& 4.7 & - & 37.7 
& - & 72.8 \\
VWFormer~\cite{yan2024multiscale} & 2024 ICLR & MiT-B0
& 3.7 & 5.8 & 38.9 
& - & 77.2\\
FeedFormer~\cite{shim2023feedformer}   & 2023 AAAI  & MiT-B0  & 4.5  & 7.8  & 39.2   
& 107.4  & 77.9   \\
SDPT~\cite{cao2024sdpt} & 2024 TITS & SDPT-Tiny 
& 3.6 & 5.7 & 39.4 
&63.4 & 77.3\\
U-MixFormer~\cite{yeom2023u}   & 2023 arxiv  & MiT-B0  & 6.1  & 6.1  & 41.2  
& 101.7  & 79.0   \\
\rowcolor{gray!20}\textbf{MacFormer (Ours)}  & \textbf{2024}   & \textbf{MiT-B0}   & \textbf{8.4}  & \textbf{6.5}   & \textbf{42.6}   & \textbf{126.0}  & \textbf{80.1}\\ 
\midrule

SegFormer~\cite{xie2021segformer}    & 2021 NeurIPS  & LVT & 3.9  & 10.6  & 39.3   
& 140.9 & 77.6   \\
FeedFormer~\cite{shim2023feedformer}   & 2023 AAAI & LVT & 4.6  & 10.0  & 41.0   
& 124.6  & 78.6  \\
U-MixFormer~\cite{yeom2023u}   & 2023 arxiv   & LVT  & 6.5  & 9.1   & 43.7   
& 122.1  & 79.9   \\
\rowcolor{gray!20}\textbf{MacFormer (Ours)}  & \textbf{2024}   & \textbf{LVT}   & \textbf{9.1}   & \textbf{10.4} & \textbf{45.9}  & \textbf{171.0}  & \textbf{81.0}\\ 
\midrule

SegNeXt~\cite{guo2022segnext}     & 2022 NeurIPS   & MSCAN-T  & 4.3   & 6.6   & 41.1  & 56.0  & 79.8   
\\
U-MixFormer~\cite{yeom2023u}  & 2023 arxiv  & MSCAN-T   & 6.7   & 7.6   & 44.4 & 90.0  & 81.0   
\\
\rowcolor{gray!20}\textbf{MacFormer (Ours)}  & \textbf{2024}   & \textbf{MSCAN-T}   & \textbf{9.0}  & \textbf{7.9}  & \textbf{46.7}  & \textbf{130.0}  & \textbf{82.0}\\ 
\bottomrule
\end{tabular}
}
\label{Table 1}
\end{center}
\end{table*}

\begin{table*}[!t]
\caption{Performance comparison with SOTA medium-weight models on ADE20K and Cityscapes.}
\begin{center}	
\scalebox{1.0}{
\begin{tabular}{llcccccc}
\toprule
\multirow{2}{*}{Method} & \multirow{2}{*}{Year} & \multirow{2}{*}{Backbone} & \multirow{2}{*}{Params. (M)$\downarrow$} & \multicolumn{2}{c}{ADE20K} & \multicolumn{2}{c}{Cityscapes} \\
&   &   &   & GFLOPs$\downarrow$  & mIoU (\%)$\uparrow$  & GFLOPs$\downarrow$  & mIoU (\%)$\uparrow$      \\ \midrule

CCNet~\cite{huang2019ccnet}   & 2019 ICCV   & ResNet-101  & 68.9  & 278.4  & 43.7   & 2224.8   & 79.5    \\
EncNet~\cite{zhang2018context}  & 2018 CVPR  &  ResNet-101 & 55.1 
& 218.8 & 44.7 & 1748.0 & 76.9\\
DeepLab-V3+~\cite{chen2018encoder}& 2018 ECCV & ResNet-101  & 52.7  & 255.1 & 44.1 & 2032.3  & 80.9  \\
Mask2Former~\cite{cheng2022masked}   & 2022 CVPR  & ResNet-101 & 63.0 & 90.0 & 47.8  & - & -   \\
Auto-DeepLab~\cite{liu2019auto}    & 2019 CVPR & Auto-DeepLab-L  & 44.4   & -  & -   & 695.0  & 80.3   \\
OCRNet~\cite{yuan2020object}     & 2020 ECCV   & HRNet-W48   & 70.5  & 164.8  & 45.6   & 1296.8  & 81.1   \\
\midrule

SegFormer~\cite{xie2021segformer}   & 2021 NeurIPS  & MiT-B1 & 13.7 & 15.9  & 42.2   & 243.7  & 78.5  \\
SegDformer~\cite{shi2022transformer} & 2022 ECCV & MiT-B1 & 14.4 & - & 44.1 
& - & - \\
SeaFormer~\cite{wanseaformer} & 2023 ICLR 2023 & SeaFormer-B & 8.6 & 1.8 & 40.2 & - & 77.7\\
SCTNet~\cite{xu2024sctnet} & 2024 AAAI & SCTNet-B & 17.4 & - & 43.0 & - & 79.8 \\
VWFormer~\cite{yan2024multiscale} & 2024 ICLR & MiT-B1 &13.7 & 13.2 & 43.2 & - & 79.0 \\

U-MixFormer~\cite{yeom2023u}   & 2023 arxiv  & MiT-B1  & 24.0  & 17.8   & 45.2 & 246.8  & 79.9  \\
SDPT~\cite{cao2024sdpt} & 2024 TITS & SDPT-Small & 11.9 & 12.7 & 46.0 &- & - \\
EfficientMod~\cite{ma2024efficient} & 2024 ICLR &EfficientMod-s & 16.7 & 28.1 & 46.0 &- & - \\

\rowcolor{gray!20}\textbf{MacFormer (Ours)}  & \textbf{2024}   & \textbf{MiT-B1}   & \textbf{33.4} & \textbf{19.2} & \textbf{47.9}  & \textbf{304.0}   & \textbf{80.9}\\ 
\midrule

SegNeXt~\cite{guo2022segnext}    & 2022 NeurIPS   & MSCAN-S  & 13.9 & 15.9  & 44.3   
& 124.6  & 81.3  \\
U-MixFormer~\cite{yeom2023u}   & 2023 arxiv  & MSCAN-S  & 24.3  & 20.8  & 48.4 & 154.0  & 81.8 \\
\rowcolor{gray!20}\textbf{MacFormer (Ours)}  & \textbf{2024}   & \textbf{MSCAN-S}   & \textbf{33.7}  & \textbf{22.1}  & \textbf{49.3}   & \textbf{211.0}  & \textbf{82.9}\\ 
\midrule

SegFormer~\cite{xie2021segformer}     & 2021 NeurIPS   & MiT-B2  & 27.5  & 62.4  & 46.5  & 717.1  & 81.0  \\
FeedFormer~\cite{shim2023feedformer}   & 2023 AAAI   & MiT-B2   & 29.1   & 42.7  & 48.0
& 522.7  & 81.5  \\
SegDformer~\cite{shi2022transformer} & 2022 ECCV & MiT-B2 & 27.6 & - & 47.5 & -&-\\
U-MixFormer~\cite{yeom2023u} & 2023 arxiv  & MiT-B2   & 35.8   & 40.0  & 48.2 & 515.0  & 81.7   \\
VWFormer~\cite{yan2024multiscale} & 2024 ICLR & MiT-B2
& 27.4 & 46.6 & 48.1 & - & -\\

\rowcolor{gray!20}\textbf{MacFormer (Ours)}  & \textbf{2024}   & \textbf{MiT-B2}   & \textbf{45.3} & \textbf{43.3} & \textbf{49.0}  & -  &-\\ 

\bottomrule
\end{tabular}
}
\label{Table 2}
\end{center}
\end{table*}

\begin{table*}[!t]
\caption{Performance comparison with SOTA heavy-weight models on ADE20K and Cityscapes.}
\begin{center}	
\scalebox{1.0}{
\begin{tabular}{llcccccc}
\toprule
\multirow{2}{*}{Method} & \multirow{2}{*}{Year} & \multirow{2}{*}{Backbone} & \multirow{2}{*}{Params. (M)$\downarrow$} & \multicolumn{2}{c}{ADE20K} & \multicolumn{2}{c}{Cityscapes} \\
&   &   &   & GFLOPs$\downarrow$  & mIoU (\%)$\uparrow$  & GFLOPs$\downarrow$  & mIoU (\%)$\uparrow$      \\ \midrule
Seg-B-Mask/16~\cite{strudel2021segmenter} & 2021 ICCV   & ViT-Base  & 106.0  & -  & 48.5  & -  & -   \\
MaskFormer~\cite{cheng2021per} & 2021 NeurIPS   & Swin-S  & 63.0  & 79.0  & 49.8  & -  & -   \\
SETR~\cite{zheng2021rethinking} & 2021 CVPR   & ViT-Large  & 318.3  & -  & 50.2  & -  & 82.2   \\
\midrule
SegNeXt~\cite{guo2022segnext}     & 2022 NeurIPS  & MSCAN-B  & 27.6  & 34.9  & 48.5  & 275.7  & 82.6   \\
\rowcolor{gray!20}\textbf{MacFormer (Ours)}  & \textbf{2024}   & \textbf{MSCAN-B}   & \textbf{46.5} & \textbf{35.3} & \textbf{49.8}  & \textbf{316.0}  & \textbf{83.8}\\ 
\midrule

SegFormer~\cite{xie2021segformer}   & 2021 NeurIPS  & MiT-B3  & 47.3   & 79.0   & 49.4 &962.9 & 81.7\\
U-MixFormer~\cite{yeom2023u} &  2023 arxiv  & MiT-B3  & 55.7       & 56.8   & 49.8 &-&-\\
VWFormer~\cite{yan2024multiscale} & 2024 ICLR & MiT-B3 & 47.3 & 63.3 & 50.3 & - & 82.4\\
\rowcolor{gray!20}\textbf{MacFormer (Ours)}  & \textbf{2024}   & \textbf{MiT-B3}   & \textbf{65.2} & \textbf{60.0}  & \textbf{50.5}   & -  &-\\ 
\midrule

SegFormer~\cite{xie2021segformer}   & 2021 NeurIPS          & MiT-B4  & 64.1       & 95.7   & 50.3 &1240.6 &82.3\\
U-MixFormer~\cite{yeom2023u} & 2023 arxiv          & MiT-B4  & 72.4       & 73.4   & 50.4 &-&-\\
VWFormer~\cite{yan2024multiscale} & 2024 ICLR & MiT-B4 & 64.0 & 79.9 & 50.8 & - & 82.7\\




\rowcolor{gray!20}\textbf{MacFormer (Ours)}  & \textbf{2024}   & \textbf{MiT-B4}   & \textbf{82.0} & \textbf{76.7}  & \textbf{50.9}   & -  &-\\
\midrule

SegFormer~\cite{xie2021segformer}   & 2021 NeurIPS          & MiT-B5  & 84.7       & 183.3  & 51.0 &1460.4&82.4\\
U-MixFormer~\cite{yeom2023u} & 2023 arxiv          & MiT-B5  & 93.0       & 149.5  & 51.9 &-&-\\
ViT-CoMer~\cite{xia2024vit}  & 2024 CVPR & ViT-CoMer-B & 144.7    & -     & 48.8 &- &- \\
ViT-CoMer~\cite{xia2024vit}  & 2024 CVPR & ViT-CoMer-L & 383.4    & -     & 54.3 &- &- \\
\rowcolor{gray!20}\textbf{MacFormer (Ours)}  & \textbf{2024}   & \textbf{MiT-B5}   & \textbf{103.0}  & \textbf{152.4} &\textbf{52.8}   & -  &-\\
\bottomrule
\end{tabular}
}
\label{Table 3}
\end{center}
\end{table*}

\section{Experiments}

To showcase the superior performance 
of the proposed MacFormer, we conducted comprehensive experiments comparing our approach with state-of-the-art methods on Cityscapes and ADE20K, two benchmark datasets commonly utilized in semantic segmentation.  

\subsection{Experiment Settings}
\subsubsection{Datasets}

\textbf{Cityscapes}~\cite{cordts2016cityscapes} comprises an extensive collection of street scenes captured in 50 European cities. These images are sized at $2048 \times 1024$.  
For this study, we specifically utilized a subset of 5,000 finely annotated images with 19 categories, divided into 2,975 images for training, 500 for validation, and 1,525 for testing purposes. 
\textbf{ADE20K}~\cite{zhou2017scene} is an impressive dataset containing 150 diverse semantic classes. It includes a total of 20,210 training images, 2,000 for validation, and an additional 3,352 for the test set. 

\subsubsection{Implementation Details}
\label{sec42}

Our experiments are 
conducted using the open-source codebase \textit{mmsegmentation}~\cite{mmseg2020} with two NVIDIA L40 GPUs. 
We employ standard data augmentation techniques, including random horizontal flipping, random scaling (0.5 to 2), and random cropping to $512 \times 512$ resolution for ADE20K and $1024 \times 1024$ for Cityscapes. 
We utilize the AdamW optimizer~\cite{loshchilov2017decoupled} to train our models, initializing with a learning rate of $6 \times 10^{-5}$. 
A poly-learning rate decay policy is applied, and the models are trained for 160K iterations on both the ADE20K and Cityscapes datasets. 

Furthermore, it is important to highlight that MacFormer is backbone-agnostic, allowing seamless integration with all existing network architectures as its backbone network. To showcase this capability and ensure a fair comparison, we employed various backbones with MacFormer, including MiT-B0 to B5, LVT, and MSCAN-T/S/B, for the comparison.

\begin{table}[!htb]
\caption{Ablation study on the effectiveness of the proposed Mutual Agent Cross-Attention (MACA) and Frequency Enhancement Modules (FEM) with 
different encoder backbones. 
In the ``Method'' column of the table, ``CA'' denotes Cross-Attention, ``MCA'' denotes Mutual Cross-Attention, ``MACA'' denotes Mutual Agent Cross-Attention, and ``FEM'' denotes Frequency Enhancement Module.}
\begin{center}	
\setlength\tabcolsep{5pt}
\scalebox{1.0}{
\begin{tabular}{lcccccc}
\toprule
\multirow{2}{*}{Method}                                  & \multirow{2}{*}{Backbone} & \multicolumn{1}{c}{\multirow{2}{*}{Params.(M)$\downarrow$}} & \multicolumn{2}{c}{ADE20K}                                            \\
&    & \multicolumn{1}{c}{}   & \multicolumn{1}{c}{GFLOPs$\downarrow$} & \multicolumn{1}{c}{mIoU (\%)$\uparrow$}  \\ \midrule

CA (Baseline)  
& MiT-B0 
& \multicolumn{1}{c}{6.1}    
& \multicolumn{1}{c}{6.1}    
& \multicolumn{1}{c}{41.2}  \\

MCA   & MiT-B0     & 7.0    &   5.6   & 41.7           \\
MACA   & MiT-B0    & 8.4   &  5.9  & 42.1            \\ 
\rowcolor{gray!20}MACA + FEM & MiT-B0   & 8.4  & 6.5      & 42.6       \\ 
\midrule

CA (Baseline) & MiT-B1    
& \multicolumn{1}{c}{24.0}      
& \multicolumn{1}{c}{17.8}    
& \multicolumn{1}{c}{45.2} 
 \\

MCA   & MiT-B1     & 27.3    &   16.3   & 46.3 \\
MACA   & MiT-B1    & 33.3   & 17.0  & 47.1    \\ 
\rowcolor{gray!20}MACA + FEM & MiT-B1   & 33.4  & 19.2   & 47.9     \\ 
\midrule 

CA (Baseline)                   
& LVT                    
& \multicolumn{1}{c}{6.5}                     & \multicolumn{1}{c}{9.1}    
& \multicolumn{1}{c}{43.7} 
\\

MCA   & LVT     & 7.4    & 8.3     & 44.5          \\
MACA   & LVT    & 9.0    & 8.8  & 45.1     \\ 
\rowcolor{gray!20}MACA + FEM & LVT   & 9.1  & 10.4     & 45.9 \\ 
\midrule 

CA (Baseline)                   
& MSCAN-T                    
& \multicolumn{1}{c}{6.7}                     & \multicolumn{1}{c}{7.6}    
& \multicolumn{1}{c}{44.4} 
\\

MCA   & MSCAN-T      & 7.5    & 7.0     & 45.0  \\
MACA   & MSCAN-T     & 9.0   & 7.3  & 45.8    \\ 
\rowcolor{gray!20} MACA + FEM & MSCAN-T    & 9.0  & 7.9    & 46.7 \\ 
\midrule

CA (Baseline)  
& MSCAN-S                    
& \multicolumn{1}{c}{24.3}                    & \multicolumn{1}{c}{20.8}   
& \multicolumn{1}{c}{48.4} 
\\

MCA   & MSCAN-S      & 27.6    & 19.3     & 48.6      
\\
MACA   & MSCAN-S     & 33.6   & 20.0  & 48.9  \\ 
\rowcolor{gray!20}MACA + FEM & MSCAN-S    & 33.7  & 22.1     & 49.3  \\ 

\bottomrule
\end{tabular}
}
\label{Table 4}
\end{center}
\end{table}

\begin{table}[tbp]
\caption{Performance comparison of varying frequency thresholds on ADE20K and Cityscapes datasets with MiT-B0 and MiT-B1 backbones.}
\begin{center}	
\renewcommand\arraystretch{1.2}
\scalebox{0.95}{
\begin{tabular}{cc|ccccc}
\toprule
\multicolumn{2}{c|}{Frequency Thresholds}        
& 0.1 & 0.2 & 0.3 & 0.4 & 0.5 \\ \midrule
\multirow{2}{*}{MiT-B0}
& ADE20K     & 41.32    & 41.98    & 42.17    & 42.35    & \cellcolor{gray!20}\textbf{42.62}    \\ \cline{2-2}
& Cityscapes & 79.53    & 79.72   & 79.86     & 80.01    & \cellcolor{gray!20}\textbf{80.13}   \\
\midrule
\multirow{2}{*}{MiT-B1}
& ADE20K     & 46.58    & 46.72    & 47.22    & 47.15    & \cellcolor{gray!20}\textbf{47.93}    \\ \cline{2-2}
& Cityscapes & 80.63    & 80.58   & 80.78     & 80.76    & \cellcolor{gray!20}\textbf{80.92}   \\

\bottomrule
\end{tabular}
}
\label{Threshold}
\end{center}
\end{table}

\subsection{Quantitative Results}

We classified state-of-the-art (SOTA) models into three tiers: light-weight, medium-weight, and heavy-weight.

\subsubsection{Light-weight Models}

Table~\ref{Table 1} displays the comparison results with SOTA light-weight models, providing parameter size, Floating Point Operations (FLOPs), and mIoU obtained on ADE20K and Cityscapes datasets. 


As indicated by the performance metrics of lightweight models in Table~\ref{Table 1}, our MacFormer (MiT-B0) model notably achieved a mIoU of 42.6\% on ADE20K, showcasing exceptional efficiency with only 8.4 million parameters and 6.5 GFLOPs. 

Under similar or even lower FLOP conditions, our results have achieved SOTA 
performance. Specifically, compared to SegFormer~\cite{xie2021segformer}, FeedFormer~\cite{shim2023feedformer}, and U-MixFormer~\cite{yeom2023u}, our model has enhanced mIoU by 5.2\%, 3.4\%, and 1.4\%, respectively. On the Cityscapes dataset, our model also exhibited significant performance improvement, achieving a mIoU of 80.1\% with only 126.0 GFLOPs. This marks a 3.9\% increase in mIoU while maintaining nearly identical computational complexity as SegFormer~\cite{xie2021segformer}. 

Similarly, utilizing the encoder backbones LVT and MSCAN-T, our model delivers notably higher mIoU results compared to other models. Specifically, on the ADE20K dataset, 
our MacFormer model based on MSCAN-T achieves a performance gain of 2.3\%, while requiring only an additional 0.3 GFLOPs compared to U-MixFormer~\cite{yeom2023u}.


\subsubsection{Medium-weight Models} 
Table~\ref{Table 2} shows that using MiT-B1 as the backbone, our model has achieved 
47.9\% mIoU on ADE20K and an impressive 80.9\% mIoU on Cityscapes with only 33.4 million parameters. In comparison, Mask2Former~\cite{cheng2022masked} requires 63.0 million parameters to achieve similar performance, doubling our model's size and nearly quintupling the GFLOPs to 90.0. Similarly, DeepLab-V3+~\cite{chen2018encoder}, achieving 80.9\% mIoU on Cityscapes, requires 6.7 times the GFLOPs of our model, showcasing the fine balance between performance and efficiency of our approach. Moreover, when employing MiT-B2 and MSCAN-S as encoder backbones, our model consistently outperforms competitors by at least 0.8\% on the ADE20K dataset. When using the MiT-B2 backbone, our model exhibits a substantial 2.5\% mIoU improvement over Segformer~\cite{xie2021segformer}. Furthermore, with MSCAN-S, it surpasses Segformer by 5\% and U-mixFormer~\cite{yeom2023u} by 0.9\%. Although the superiority margin is less pronounced on the Cityscapes dataset, our model still demonstrates a  performance gain of more than 
1\%, 
signifying the effectiveness of our proposed approach.

\subsubsection{Heavy-weight Models} 
Table~\ref{Table 3} demonstrates that the proposed MacFormer outperforms state-of-the-art heavy-weight models when combined with equivalent medium-sized encoders such as MSCAN-B, MiT-B3, MiT-B4, and MiT-B5. When coupled with MSCAN-B as the backbone for SegNeXt~\cite{guo2022segnext}, our model achieves a significant 1.3\% mIoU improvement at a marginal 0.4 GFLOPs cost. Comparisons with MiT-B3/4/5 are omitted due to practical constraints in training under the same batch size conditions for Cityscapes' high resolution. However, on the ADE20K dataset, our method has demonstrated a superiority of over 0.5\% compared to others, highlighting the substantial superiority of our approach.

\begin{figure*}[htbp]
	\centerline{\includegraphics[width=16cm]{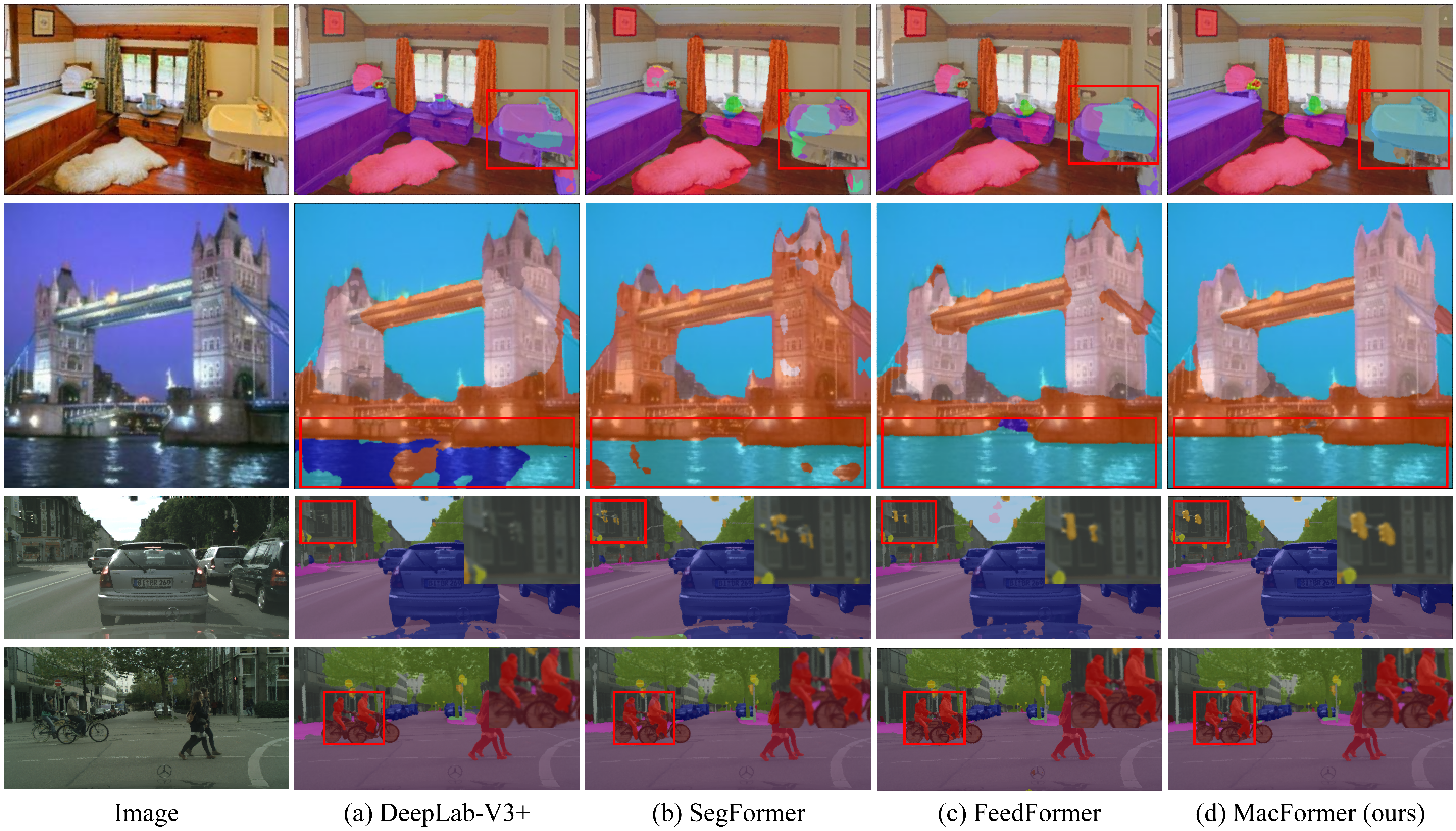}}
	\caption{Qualitative comparison with DeepLab-V3+~\cite{chen2018encoder}, SegFormer~\cite{xie2021segformer}, and Feedformer~\cite{shim2023feedformer} on ADE20K and Cityscapes.}
	\label{Figure 6}
\end{figure*}

\begin{table*}[htbp]
  \centering
  \caption{Ablation study of deploying the proposed Frequency Enhancement Module (FEM) in different networks.}
  \begin{subtable}{0.5\linewidth}
    \centering
    \caption{Ablation study on FEM in SegFormer~\cite{xie2021segformer} and SegNeXt~\cite{guo2022segnext}.}
    \begin{tabular}{llccc}
    \toprule
\multirow{2}{*}{Method}    & \multirow{2}{*}{Backbone} & \multirow{2}{*}{FEM} & \multicolumn{2}{c}{mIoU (\%)$\uparrow$} \\
&   &    & ADE20K      & Cityscapes      \\ \midrule

\multirow{4}{*}{SegFormer~\cite{xie2021segformer}}   
& \multirow{2}{*}{MiT-B0}                       
& \multicolumn{1}{c}{\XSolidBrush}      
& \multicolumn{1}{c}{37.4} 
& \multicolumn{1}{c}{76.2}    \\
&  & \Checkmark     &  \cellcolor{gray!20}38.4 (+1.0)  
& \cellcolor{gray!20}76.9 (+0.7)     \\           
& \multirow{2}{*}{MiT-B1}   & \XSolidBrush   & 42.2     & 78.5     \\
&   & \Checkmark    &   \cellcolor{gray!20}42.9 (+0.7)       
& \cellcolor{gray!20}78.9 (+0.4)      \\
\midrule
\multirow{4}{*}{SegNeXt~\cite{guo2022segnext}}  
& \multirow{2}{*}{MSCAN-T}                      
& \multicolumn{1}{c}{\XSolidBrush}          
& \multicolumn{1}{c}{41.1}    
& \multicolumn{1}{c}{79.8}     \\
&    & \Checkmark   &  \cellcolor{gray!20}42.0 (+0.9)     
&  \cellcolor{gray!20}80.1 (+0.2)       \\
& \multirow{2}{*}{MSCAN-S}   & \XSolidBrush   
& 44.3   & 81.3  \\
&  & \Checkmark  & \cellcolor{gray!20}45.1 (+0.8) 
& \cellcolor{gray!20}81.8 (+0.5) \\ 
\bottomrule
\end{tabular}
    \label{subtab1}
  \end{subtable}%
  \begin{subtable}{0.5\linewidth}
    \centering
    \caption{Ablation study on FEM in FeedFormer~\cite{shim2023feedformer} and U-mixFormer~\cite{yeom2023u}.}
    \begin{tabular}{llccc}
    \toprule
\multirow{2}{*}{Method}    & \multirow{2}{*}{Backbone} & \multirow{2}{*}{FEM} & \multicolumn{2}{c}{mIoU (\%)$\uparrow$} \\
&   &    & ADE20K      & Cityscapes      \\ \midrule

\multirow{4}{*}{FeedFormer~\cite{shim2023feedformer}}  
& \multirow{2}{*}{MiT-B0}   
& \multicolumn{1}{c}{\XSolidBrush}  
& \multicolumn{1}{c}{39.2}    
& \multicolumn{1}{c}{77.9}     \\
&       & \Checkmark     & \cellcolor{gray!20}40.5(+1.3)    
& \cellcolor{gray!20}78.2 (+0.3)    \\
& \multirow{2}{*}{MiT-B2}  & \XSolidBrush  & 48.0  & 81.5   \\
&  & \Checkmark  & \cellcolor{gray!20}48.2 (+0.2)   
& \cellcolor{gray!20}81.9 (+0.4)   \\
\midrule

\multirow{4}{*}{U-mixFormer~\cite{yeom2023u}} 
& \multirow{2}{*}{MiT-B0}  & \multicolumn{1}{c}{\XSolidBrush}            
& \multicolumn{1}{c}{41.2}    
& \multicolumn{1}{c}{79.0}     \\
& & \Checkmark & \cellcolor{gray!20}41.7 (+0.5) 
& \cellcolor{gray!20}79.4 (+0.4)  \\

& \multirow{2}{*}{MiT-B1} & \multicolumn{1}{c}{\XSolidBrush}          
& \multicolumn{1}{c}{45.2}    
& \multicolumn{1}{c}{79.9}   \\
&     & \Checkmark   & \cellcolor{gray!20}45.8 (+0.6)    
& \cellcolor{gray!20}80.2 (+0.3)   \\
      \bottomrule
    \end{tabular}
    \label{subtab2}
  \end{subtable}
  \label{Table 5}
\end{table*}

\subsection{Qualitative Results}

Fig.~\ref{Figure 6} depicts the qualitative comparison of segmentation outcomes generated by our MacFormer and several prominent methods (such as DeepLab-V3+~\cite{chen2018encoder}, SegFormer~\cite{xie2021segformer}, and Feedformer~\cite{shim2023feedformer}) on the ADE20K and Cityscapes datasets. 

The comparison 
reveals that our approach achieves more precise segmentation at object boundaries, excelling in two aspects. Firstly, it ensures accurate segmentation of identical objects, unaffected by nearby large 
objects. For example, when comparing the predictions in the first and second rows, the sink is precisely classified without interference from adjacent furniture like bedside tables or beds. Additionally, the river surface is accurately classified without being influenced by shadows. This accuracy is attributed to the rich and precise semantic feature representation obtained from multi-scale encoding and the interdependencies among long-distance pixels in our proposed MACA model. Secondly, our method demonstrates superior accuracy in segmenting small objects. As seen in the third and fourth rows, small targets such as streetlights and specific body parts of cyclists (\textit{e.g.}, buttocks and feet) are classified with precision. This enhancement is a result of the comprehensive supplementation of detailed information, enabling robust handling of challenging areas and the edges of small objects.

\subsection{Ablation Studies}

We further conducted a series of ablation studies to validate the effectiveness of the two proposed modules, i.e., MACA and FEM, and their influence on segmentation performance across different backbone networks.

\subsubsection{Effectiveness of the MACA and FEM}
In Table~\ref{Table 4}, we systematically analyze the performance of our proposed model across various encoder backbones. 
The experimental settings were kept consistent across all trials to ensure experimental fairness. 

When comparing the baseline CA with MCA across different encoder backbones, a notable performance improvement was observed. This enhancement, marked by mutual feature supplementation, was evident in both the ADE20K and Cityscapes datasets. However, this improvement is accompanied by an increase in parameters and FLOPs, reflecting the inherent computational complexity of self-attention. 

Introducing agent tokens in MACA introduces controllable and learnable parameters into the attention mechanism, with the computational load adjustable by the dimension of agent tokens. This enhancement results in a notable improvement of around 0.5\% mIoU on the Cityscapes dataset and a minimum of 0.3\% mIoU on ADE20K. 

Finally, we validated the effectiveness of the Frequency Enhancement Module (FEM), resulting in a consistent performance improvement in segmentation with a minimal increase in parameters and computational complexity. This validation was supported by a performance gain of 0.4\% to 0.9\% mIoU on the ADE20K dataset and an average increase of 0.3\% mIoU on the Cityscapes dataset.

\subsubsection {Impact 
of Varying Frequency Thresholds on Performance}
In our FEM, we extracted both high- and low-frequency components 
from the Fourier-transformed features using threshold truncation, similar to high-pass and low-pass filtering. The threshold setting had a substantial impact on the results. 
Table~\ref{Threshold} displays various experimental results under different threshold settings, employing MiT-B0 and MiT-B1 backbones. 
It can be observed that our network achieves optimal performance when the threshold is set to 0.5.

\subsection {Further Study on the Effectiveness of FEM.}
Furthermore, to demonstrate the effectiveness and portability of the proposed Frequency Enhancement Module (FEM), we integrated this module into other models to assess its performance, as illustrated in Table~\ref{Table 5}. The incorporation of FEM resulted in significant improvements across four different methodologies. On the ADE20K dataset, the improvement peaked at 1.3\%, while on the Cityscapes, it achieves a maximum increase of 0.7\%. Even the smallest enhancement, at 0.2\% for both datasets, emphasizes the substantial benefits brought by integrating the frequency domain.

\subsection{Comparison with Mamba Series Approaches}
We have conducted 
experimental comparisons with some of the latest works from 2024 based on the hot research backbone of Mamba~\cite{gu2023mamba,zhu2024vision}, including PlainMamba~\cite{yang2024plainmamba}, LocalVMamba~\cite{huang2024localmamba} and
VMamba~\cite{liu2024vmamba}, shown in Table~\ref{vs_2024}. As can be seen, our MacFormer 
achieves a better balance between performance and efficiency.

\begin{table}[t]
\centering
  \caption{
  Comparisons with the latest Mamba-based methods on the ADE20K dataset (input size: $512 \times 2048$).}
\setlength\tabcolsep{1.8pt}
\renewcommand\arraystretch{1.2}
\scalebox{1.0}{
\begin{tabular}{lcccc}
\toprule
Method  & Backbone & Param. (M)$\downarrow$ & GFLOPs$\downarrow$ & mIoU (\%)$\uparrow$ \\ \midrule
PlainMamba~\cite{yang2024plainmamba} & PlainMamba-L3 & 81.0 & - & 49.1\\
LocalVMamba~\cite{huang2024localmamba}  & LocalVMamba-S & \textbf{81.0} & 1095.0 & 51.0\\
VMamba~\cite{liu2024vmamba}  & VMamba-B & 122.0 & 1170.0 & 51.6  \\
\rowcolor{gray!20}\textbf{MacFormer (ours)}      & MiT-B4       & \textbf{82.0}  & \textbf{358.0} & 50.9  \\
\bottomrule
\end{tabular}
}
\label{vs_2024}
\end{table}

\begin{table}[t]
    \centering
    \caption{Comparison with the Frequency-Adaptive Dilated Convolution (FADC) approach~\cite{chen2024frequency} on ADE20K.}
    \renewcommand\arraystretch{1.2}
    \scalebox{1.0}{
        \begin{tabular}{lcccc}
            \toprule
            Method    & Backbone & Param. (M)$\downarrow$ & GFLOPs$\downarrow$ & mIoU (\%)$\uparrow$ \\ \hline
            FADC~\cite{chen2024frequency}      & ResNet-50 & 67.0 & 949.0 & 45.5 \\
            \rowcolor{gray!20}\textbf{MacFormer} & LVT & \textbf{9.1} & \textbf{59.8} & \textbf{45.9} \\
            \midrule
            FADC~\cite{chen2024frequency}      & HorNet-B & 128.0     & 1176.0    & 51.5  \\ 
            \rowcolor{gray!20}\textbf{MacFormer} & MiT-B5   & \textbf{103.0}     & \textbf{442.0}     & \textbf{52.8}  \\
            \bottomrule
        \end{tabular}
    }
\label{vs_FADC}
\vspace{-1.3em}
\end{table}

\subsection{Comparison with the Frequency Domain Method}

Although there are frequency domain modules reported for other vision tasks, 
frequency has been rarely explored in semantic segmentation. 
Also, we intend to address boundary issues by cleverly combining the high and low-frequency characteristics exhibited in frequency domains, instead of rigidly applying frequency information to segmentation. 
To demonstrate the efficacy of our approach, we compared our MacFormer with the recent Frequency-Adaptive Dilated Convolution (FADC)~\cite{chen2024frequency} approach, which adaptively set dilated convolution rates based on frequency information. The comparison results, shown in Table~\ref{vs_FADC}, indicate that our model is a better choice for resource-limited application scenarios.

\begin{table*}[!t]
\caption{Inference speed comparisons on Cityscapes images 
(image resolution: $2048 \times 1024$).}
\begin{center}
\renewcommand\arraystretch{1.2}
\scalebox{1.0}{
\begin{tabular}{llccccc}
\toprule
Method  & Year & Backbone & Params. (M)$\downarrow$ & GFLOPs$\downarrow$ & mIoU (\%)$\uparrow$ & Inference (ms)$\downarrow$ \\
\midrule
PSPNet~\cite{zhao2017pyramid} & 2017 CVPR & MobileNet-V2     & 13.7   & 423.4   & 70.2  & \textbf{26.7}   \\

DeepLab-V3+~\cite{chen2018encoder} & 2018 ECCV & MobileNet-V2   & 15.4  &  125.5      & 75.2     & 36.0 \\

SegFormer~\cite{xie2021segformer}& 2021 NeurIPS & MiT-B0 & \textbf{3.8}   & 125.5  & 76.2     & 44.8       \\

FeedFormer~\cite{shim2023feedformer}& 2023 AAAI  & MiT-B0    & 4.5 & 107.4       & 77.9     & 54.4       \\

U-mixFormer~\cite{yeom2023u} & 2023 arxiv & MiT-B0      & 6.1    &  \textbf{101.7}      & 79.0     & 55.4       \\
\midrule
\rowcolor{gray!20} \textbf{MacFormer(ours)} & 2024 & MiT-B0  & 8.4     & 126.0       & \textbf{80.1}     & 60.3    \\
\bottomrule
\end{tabular}
}
\label{Table 6}
\end{center}
\end{table*}

\subsection{Limitation and Future Work}
While our MacFormer model has demonstrated competitive results in terms of computational efficiency and mIoU, it does exhibit slightly higher FLOPs when processing large images, as indicated in Table~\ref{Table 6}. This can be attributed to the inherent structure of UNet, especially its lateral connections and multi-scale fusion in the decoder section, both of which contribute to the network's computational load. To address this, future work could explore techniques such as knowledge distillation to transfer the heavy encoder components from the teacher network to the student network, thereby reducing the burden and enhancing inference speed.

\section{Conclusion}
In this paper, we introduced an effective semantic segmentation solution, MacFormer, which integrates two innovative modules to enhance features in the decoder across spatial and frequency domains. In the spatial domain, we implemented a mutual agent cross-attention mechanism to facilitate interaction between shallow and deep features, utilizing agent tokens to manage computational complexity. In the frequency domain, we utilized high and low-frequency information to complement each other, addressing the issue of insufficient structural information from deep layers due to the loss of boundary details. Experimental results on benchmark datasets have shown that MacFormer delivers exceptional performance in segmentation accuracy and efficiency, particularly in object boundary areas.

\bibliographystyle{IEEEtran}
\bibliography{reference}


 




\vfill

\end{document}